\documentclass[runningheads]{llncs}
\usepackage[T1]{fontenc}
\usepackage{graphicx}
\usepackage{adjustbox}
\usepackage{array}
\newcolumntype{R}[2]{%
    >{\adjustbox{angle=#1,lap=\width-(#2)}\bgroup}%
    l%
    <{\egroup}%
}
\newsavebox\CBox 
\def\textBF#1{\sbox\CBox{#1}\resizebox{\wd\CBox}{\ht\CBox}{\textbf{#1}}}
\usepackage{latexsym}
\usepackage[T1]{fontenc}
\usepackage[utf8]{inputenc}
\usepackage{graphicx}
\usepackage{ragged2e}
\usepackage{multirow}
\usepackage{svg}
\usepackage{float}
\usepackage{booktabs}
\usepackage{hyperref}
\usepackage{epstopdf,epsfig}
\usepackage{amsmath}

\usepackage[font=small,labelfont=bf]{caption}

\newcommand{\y}{\mathbf{y}}
\newcommand{\z}{\mathbf{z}}
\newcommand{\cc}{\mathbf{c}}
\usepackage{dblfloatfix}
\newcommand{\thickhline}{%
    \noalign {\ifnum 0=`}\fi \hrule height 1pt
    \futurelet \reserved@a \@xhline
}
\usepackage{amsfonts} 
\usepackage{microtype}

\newcommand\scalemath[2]{\scalebox{#1}{\mbox{\ensuremath{\displaystyle #2}}}}
\newcommand*\rot{\multicolumn{1}{R{90}{-0.8em}}}

\begin{document}
\title{Knowledge Graph Embeddings for Multi-Lingual Structured Representations of Radiology Reports}
\titlerunning{Knowledge Graph Embeddings for Radiology Reports Across Languages}
%
\author{Tom van Sonsbeek  \and
Xiantong Zhen \and
Marcel Worring}
\authorrunning{T van Sonsbeek et al.}
%
\institute{University of Amsterdam, The Netherlands\\
\email{\{t.j.vansonsbeek, x.zhen, m.worring\}@uva.nl}}
\maketitle              
\begin{abstract}
The way we analyse clinical texts has undergone major changes over the last years. The introduction of language models such as BERT led to adaptations for the (bio)medical domain like PubMedBERT and ClinicalBERT. These models rely on large databases of archived medical documents. While performing well in terms of accuracy, both the lack of interpretability and limitations to transfer across languages limit their use in clinical setting.
We introduce a novel light-weight graph-based embedding method specifically catering radiology reports. It takes into account the structure and composition of the report, while also connecting medical terms in the report through the multi-lingual SNOMED Clinical Terms knowledge base. The resulting graph embedding uncovers the underlying relationships among clinical terms, achieving a representation that is better understandable for clinicians and clinically more accurate, without reliance on large pre-training datasets. We show the use of this embedding on two tasks namely disease classification of X-ray reports and image classification. For disease classification our model is competitive with its BERT-based counterparts, while being magnitudes smaller in size and training data requirements. For image classification, we show the effectiveness of the graph embedding leveraging cross-modal knowledge transfer and show how this method is usable across different languages.
\end{abstract}
\section{Introduction}

Processing of medical text underwent major changes with the emergence of Transformer-based models. Fine-tuned versions of the Bidirectional Encoder Representation of Transformers (BERT) \cite{devlin2018bert} model, such as ClinicalBERT \cite{alsentzer2019clinicalBERT} and BioBERT \cite{lee2020biobert} are highly effective \cite{gu2021domain}. In particular ClinicalBERT works well in tasks regarding radiology reports of X-ray scans, such as text based disease classification and report generation \cite{casey2021systematic}. However, there are crucial complications of directly applying general NLP methods to the medical domain making a case for creating more domain-specific solutions to processing of medical text: (1) BERT produces embeddings that are computationally expensive and data-inefficient. Since there is no prior content knowledge, high parameter count and deep models are needed to optimally model text. To bridge the gap between generic text and medical text there is a dependency on fine-tuning with large medical text datasets. (2) BERT and its fine-tuned versions are self-supervised methods. Effective connections and patterns are learnt, but these certainly are not equivalent to medical knowledge. The lack of explainability and intuition behind them complicates their usage in synergy with clinicians, who need to understand how these models work before they can trust them \cite{litjens2017survey}. (3) BERT models largely focus on English language.  These large language models can not be adapted for multi-lingual use with the same effectiveness. Spanish equivalents of BERT (BETO~\cite{CaneteCFP2020}) and ClinicalBERT (Bio-cli-52k~\cite{carrino2021biomedical}) are trained with around ten times less data and hence have lower performance. 

As an alternative to self-supervision we can use formalized medical knowledge. The Universal Medical Language System (UMLS) contains standardized definitions and relationships within medical terminologies and vocabularies across 25 languages~\cite{bodenreider2021domain}. Examples are ontologies for primary care (ICPC), genes (GO), clinical terms (SNOMED CT), drugs (Rxnorm) and even billing (ICD10). The UMLS can be used among various national hospitals, but also across countries. This can be especially beneficial to countries that do not have access to large medical datasets, due to their smaller population size or lack of financial resources. 
 
The specific ontologies within the UMLS can provide additional advantages. SNOMED CT provides relationships between concepts within their respective ontology. The added information from this knowledge base can be beneficial, since expert-level annotation is not in abundance in the medical domain. One particularly useful application of SNOMED CT is in radiology reports, which are widely available in public datasets, but are largely un-annotated.  The structure of UMLS and SNOMED CT make them suitable for representation with knowledge graphs, which can efficiently represent structured sets of entities \cite{ji2021survey,chang2020benchmark}. 

In this paper we propose the first graph embedding based method for structured representation of radiology reports which incorporates information from existing medical knowledge by leveraging the structure and composition of the text. (i) Experiments show that the proposed graph embedding achieves competitive performance. These report embeddings are a computationally more efficient, more explainable and intuitive alternative to existing embedding methods. (ii) We show that the usage of UMLS and SNOMED CT allows for easy translation across languages. (iii) Lastly, the proposed report graph can be used in a multi-modal setting for cross-modal knowledge transfer to images, enabling improved image-based disease classification. 


\subsubsection{Related work}

There are two  methods currently used most for embedding of medical text. The first one is BioWordVec~\cite{zhang2019biowordvec}, a word2vec~\cite{mikolov2013efficient} inspired embedding pre-trained on biomedical datasets. The second is a class of methods, namely fine-tuned versions of BERT and BETO, such as BioBERT~\cite{lee2020biobert}, PubMedBERT~\cite{pubmedbert}, ClinicalBERT~\cite{alsentzer2019clinicalBERT} and Bio-cli-52k~\cite{carrino2021biomedical}. These models are fine-tuned using respectively (bio)medical and clinical datasets. These type of embeddings are currently used in most recent state-of-the-art methods which use chest X-ray reports and outperform previous methods by a large margin. A new embedding should be compared against these BERT-based methods. 

Knowledge graphs have been used to improve diagnosis based on patient records \cite{heilig2022refining}, for improvement of entity extraction from radiology reports~\cite{jain2021radgraph} and to supplement X-ray image diagnosis~\cite{prabhakar2022structured}. Numerous methods were developed for using knowledge graphs in radiology report generation. The graph to generate a report is typically a small graph ($\sim$15 nodes) containing disease labels. This proves to be an effective way to capture the global context of a report~\cite{zhang2020radiology,liu2021auto,yang2021knowledge,kale2022knowledge,liu2021exploring,hu2021word,hu2022graph,zhang2022improving,yan2022memory}. 
Our method proposes an embedding \textit{from} a radiology report, instead of \textit{for the generation of} a radiology report. There have not been prior works on structured representations encompassing full radiology reports with knowledge graphs and medical ontologies.

\section{Methodology}

\begin{figure}[!t]
    \centering
    \includegraphics[width=\linewidth]{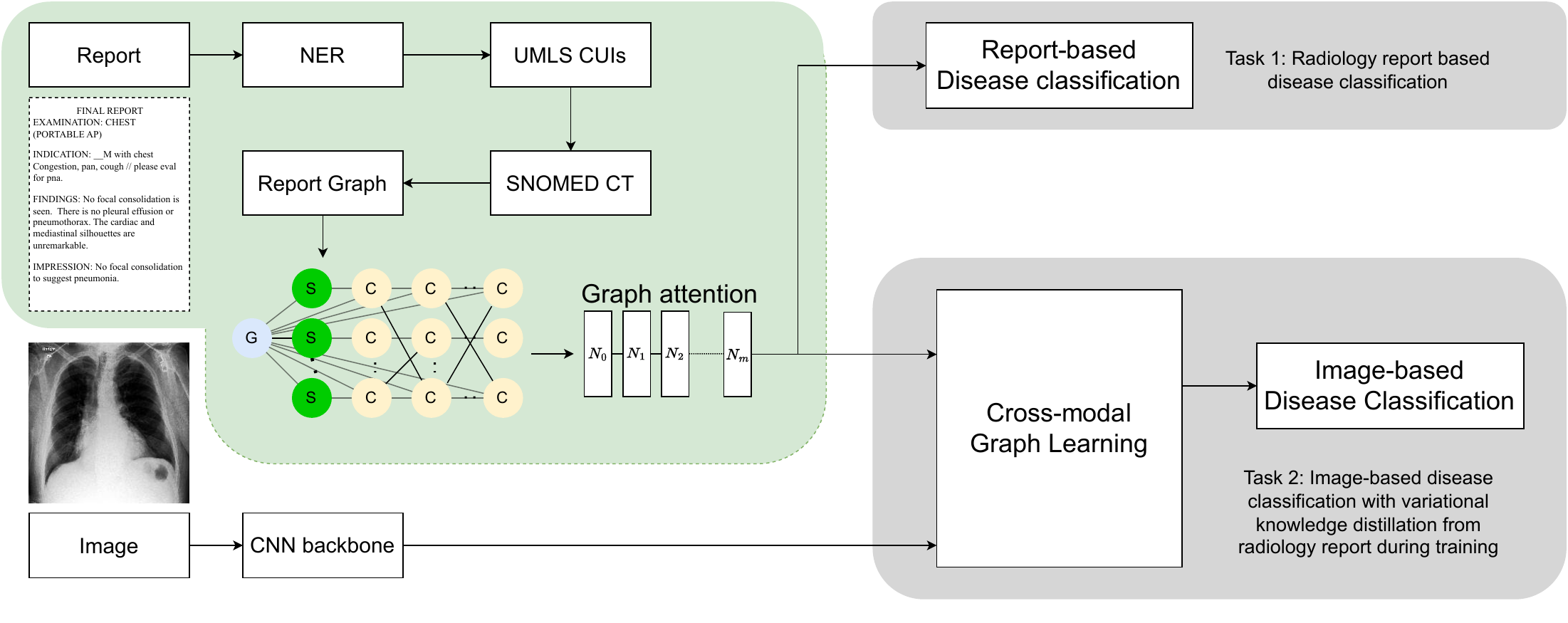}
    \caption{Schematic representation of construction and evaluation methods for knowledge graph embeddings for representations of radiology reports. We evaluate the proposed graph embeddings for text (task 1) and image (task 2) based disease classification.}
    \label{fig:eval_struc}
\end{figure}

Our proposed method consists of three components: entity extraction from the radiology report, graph construction, and graph encoding, as illustrated in \autoref{fig:eval_struc}. The nodes in the graph correspond to words in the report that match terms from clinical databases. Graph edges encode relationships between those terms and their location in the report \footnote{\url{github.com/tjvsonsbeek/knowledge_graphs_for_radiology_reports.git}}. 

\subsubsection{Named Entity Recognition}

Clinical concepts in the plain text of the radiology report $R$ are extracted using Named Entity Recognition (NER). A widely used tool for NER on English UMLS concepts is $\mathrm{MetaMap}$ \cite{aronson2010overview}. For Spanish we use UMLSMapper~\cite{perez2}. We can extract UMLS Concept Unique Identifiers (CUIs): $\{\mathbf{u}^{0},\mathbf{u}^{1},..\} = \mathrm{NER}(R)$ from $R$. The corresponding SNOMED CT concept for each CUI forms the final set of clinical concepts from report $R: \{\cc^{0},\cc^{1},..\}$.

\subsubsection{Graph Construction}
We consider undirected graph $G_R = (N_R, E_R)$ defined by a set of nodes $N_R = \{n_0, n_1, ..\}$ and edges $E_R = \{e^{l\leftrightarrow m}, ..\}$, with $e^{l\leftrightarrow m}  = (n_l, n_m)$. To capture the structure of the radiology report we consider each sentence of the report separately with sentence node $\mathbf{s}^{j}$ of sentence $j$. SNOMED CT concepts extracted from report $R$ yield a set of concepts per sentence: $\{\cc^{j,0},\cc^{j,1},..\}$. 

Global connection node $\mathbf{g}$ captures context between nodes. This gives us the following set of nodes: $N_R = \{\mathbf{s}^{j},\mathbf{g},\mathbf{c}\}$. We consider three different types of edges:
$$E_R = \{e^{c\leftrightarrow c}, e^{s\leftrightarrow c}, e^{g\leftrightarrow s},e^{g\leftrightarrow c}\} = \{(\cc^{l},\cc^{m})\},\{(\cc^{l},\mathbf{s}^{j})\},\{(\mathbf{s}^j,\mathbf{g})\},\{(\cc^{l},\mathbf{g})\}\}$$
\begin{enumerate}
    \item Edges $e^{c\leftrightarrow c}$ between concept nodes $c$ that hold a contextual connection to each other due to a corresponding relation in the SNOMED CT ontology.  
    \item Edges $e^{s\leftrightarrow c}$ from the sentence nodes $\mathbf{s}^{j}$ to all $\cc$ within their respective sentence capturing the local composition of the report. 
    \item Global connection node $\mathbf{g}$ is connected to every concept node node through edges $e^{g\leftrightarrow c}$ and $e^{g\leftrightarrow s}$ to enable interaction across the entire report.

\end{enumerate}

\subsubsection{Graph Encoding}

Graph attention networks leverage the self-attention mechanism to enable nodes in $G_R$ to efficiently attend to their neighborhoods \cite{velivckovic2018graph}. This is the current preferred method in encoding of (knowledge) graphs \cite{zhou2020graph}. We consider node $N_{R}^p$ and neighboring nodes $N_{R}^q$, with their respective weight matrices $\mathbf{W}\in\mathbb{R}^{F'\times~F}$, in which $F$ is the node feature length. The normalized attention score between those nodes can be defined as:
$$a_{pq} = softmax(LeakyRELU(\mathbf{W_{att}^t}[\mathbf{W_{N_{R}^p}}\|\mathbf{W_{N_{R}^q}}])).$$
The encoded representation of node $N_{R}^p$ is: $N_{R}^{p'} = \sigma(\sum\alpha_{pq}\cdot\mathbf{W_{N_{R}^q}})$, with nonlinearity $\sigma$. The encoding of the entire graph with $n$ stacked GAT layers can be represented as: $G_{R}^{'} = f^{n}_{GAT}(N_R, E_R)$. 

\subsubsection{Evaluation tasks}

To evaluate whether the constructed graph embeddings contain information representative of the content of the radiology report, we evaluate them on two types of tasks centered around disease classification as shown in \autoref{fig:eval_struc}. 
We first evaluate the effectiveness of our knowledge graph embedding for a diagnosis classification task based on the radiology report. We make comparisons against BERT and its bio(medical) variations. Disease classification is done on the global report. This is achieved through a max pooling operation on encoded node representations $N_{R}^{'}$, followed by a classification MLP: $f_{MLP}(\cdot)$. 
The capability of out graph embedding to transfer the information it contains across modalities is tested by deploying the embeddings in the variational knowledge distillation framework (VKD)~\cite{van2021variational}, where graph embeddings will be used to aid image-based disease classifications. In VKD, a conditional latent variable model is introduced in which information is distilled from radiology report $R$ to chest X-ray scan $I$through variational inference. This architecture is inspired by conditional variational inference (CVI)~\cite{sohn2015learning}. The evidence lower bound objective (ELBO) of CVI is composed of a reconstruction term and a Kullback-Leibler (KL) divergence term:  
$\scalemath{0.9}{\mathcal{L} = \mathbb{E}\big[\log p(\y|I,\z_I)\big] - D_{\rm{KL}}\big[q(\z)||p(\z_I|I)\big]}$
in which $\z_I$ is a latent representation of $I$, classification labels are denoted by $\y$ and prior distribution $p(\z_I|I)$. $q(\z)$ is the posterior distribution over $\mathbf{z}$, which is usually set to an isotropic Gaussian distribution $\mathcal{N}(0,I)$. In VKD the posterior is set as $q(\z_R|R)$, with $z_R$ being a latent representation of $R$. This new posterior makes it possible to distill information from $R$ to $I$ by minimizing the following KL term:
$D_{\rm{KL}}\big[q(\z_R|R)||p(\z_I|I)\big]$. Through this process we are able to transfer information from the radiology report to the medical image. During the training stage of this method both image and radiology report are required as input, while during testing only image input is required. With this method we use our report graph embeddings to improve image representations.

\section{Experimental Setup}

\subsubsection{Datasets}
The datasets used for training and evaluation are: 1) \textit{MIMIC-CXR} \cite{johnson2019mimic} consisting of $377,110$ chest X-rays (both frontal and/or sagittal views) and $227,827$ anonymized radiology reports, with disease labels generated with a rule-based labeller. 2) \textit{OpenI} \cite{demner2016preparing} with $7,470$ chest X-rays and $3,955$ anonymized reports which are similarly labelled 3) \textit{PadChest} \cite{bustos2020padchest} contains $160,000$ radiology images and Spanish reports. It contains 174 disease labels similarly extracted as above and which can be condensed in the same label space as the other datasets. 

\subsubsection{Experimental Settings}
No limit is enforced on the number of entities that can be extracted from a report with NER through $\mathrm{MetaMap}$ or on the number of edges within a graph. We adopt vectorized representations of single UMLS concepts to initialize the nodes. These were obtained by pre-training on datasets with (bio)medical data \cite{beam2019clinical}. These 200-dimensional non-contextual embeddings were inspired by word2vec and can be used in our method without additional processing steps. Embedding initializations for $\mathbf{s}^{j}$ and $\mathbf{g}$ are computed by averaging the node embeddings $\cc$ over the sentence and whole graph respectively.  



The graph attention encoder consists of one, three, six or twelve sequential graph attention layers with hidden size $512$, $1024$ or $2048$. Graph classification is done through an MLP with dimensions $\{512,256,14\}$, using cross-entropy loss. Results are reported with the AUC metric, consistent with existing benchmarks. Evaluation on VKD is done with a latent space size  of $2048$ and 12 sequential graph attention layers for encoding. Other hyper-parameter settings are directly adopted from \cite{van2021variational}. A dropout rate of 0.5 is applied to all layers of the architecture. Training is done on one Ryzen 2990WX CPU and one NVIDIA RTX 2080ti GPU, with Adam optimization using early stopping with a tolerance of 1\%.
\begin{table}[!ht]
\setlength\tabcolsep{0.5em}
\centering
\caption{Comparison of our graph-based disease classification to BERT-based methods. Our Graph MLP Small and Graph MLP Large embeddings have one and three encoder layers and encoder hidden size $512$ and $1024$ respectively.}
\resizebox{\textwidth}{!}{
\begin{tabular}{llcccccccccccccccccc}
\toprule
\multicolumn{2}{c}{} & \multicolumn{7}{c}{OpenI}& \multicolumn{7}{c}{MIMIC-CXR}& \multicolumn{4}{c}{PadChest} \\ 
 \multicolumn{2}{c}{}&  \rot{BioWordVec}&\rot{BERT}&\rot{BioBERT}&\rot{PubMedBERT}&\rot{ClinicalBERT}&\rot{Graph MLP S}&\rot{Graph MLP L}&\rot{BioWordVec}&\rot{BERT}&\rot{BioBERT}&\rot{PubMedBERT}&\rot{ClinicalBERT}&\rot{Graph MLP S}&\rot{Graph MLP L}&\rot{BETO}&\rot{Bio-cli-52k}&\rot{Graph MLP S}&\rot{Graph MLP L}\\\cmidrule(r){1-2}\cmidrule(rl){3-9}\cmidrule(rl){10-16}\cmidrule(l){17-20}
\multirow{2}{*}{\begin{tabular}[c]{@{}c@{}}Inference rate\\\hspace{5pt} (1/s) \end{tabular}}&CPU&-&17  & 17  & 18  & 18  & \textBF{41}  & \underline{34}  &-& 18  & 19  & 18  & 18  & \textBF{39}  & 33  & 23  & 23  & \textBF{46}  & 43  \\
&GPU&-&208 & 213 & 212 & 216 & \textBF{398} & 325 &-& 203 & 204 & 203 & 203 & \textBF{368} & 312 & 241 & 248 & 412 & \textBF{428}\\\cmidrule(r){1-2}\cmidrule(rl){3-9}\cmidrule(rl){10-16}\cmidrule(l){17-20}
No Finding  && .924 & .796 & .861 & .907 & \textBF{.934} & .917 & \underline{.926} &.817  &.891&.914&.923&\underline{.923}&.882&\textBF{.902}&.923&.954&\textBF{.980}&\textBF{.980}\\
Enl. cardiomed. && .859 & .715 & .784 & .854 & \textBF{.864} & .852 & \underline{.861} & .863 &.855&.866&\underline{.870}& \textBF{.966} &.818&\underline{.870}&.645&.809&\underline{.863}&\textBF{.908}\\
Cardiomegaly&& \underline{.944} & .922 & .941 & .941 & \textBF{.969} & .935 & .937 &.921&.917&\underline{.925}&.937& \textBF{.979}&.904&.920&.861&\textBF{.895}&.866&\underline{.873}\\
Lung Opacity&& .953 & .820 & .902 & .950 & \underline{.973} & .966 & \textBF{.976} &.910&.923&.936&.942&\textBF{.978}&.930&\underline{.949}&\textBF{.958}&\underline{.923}&.917&.914\\
Lung Lesion && .961 & .814 & \underline{.869} & .941 & \textBF{.971} & .945 & .968 &.913&.917&.921&\underline{.930}&\textBF{.972}&.911&.927&.801&.826&\underline{.928}&\textBF{.932}\\
Edema && \textBF{.984} & .859 & .924 & .946 & \underline{.976} & .951 & .965 &.898&916&.927&.935&\textBF{.979}&.921  &\underline{.939}&.832&.913&\underline{.939}&\textBF{.957}\\
Consolidation&& \underline{.969} & .923 & .926 & .968 & \textBF{.982} & .955 & .968 &.905&.920&.935&.942&\textBF{.979}&.958  &\underline{.965}&.701&.749&\underline{.871}&\textBF{.974}\\
Pneumonia  && \underline{.983} & .878 & .900 & .940 & \textBF{.982} & .947 & .968 &.932&.925&.939&\underline{.949}&\textBF{.962}&.940  &.945&.794&.823&\underline{.853}&\textBF{.877}\\
Atelectasis&& \underline{.981} & .847 & .897 & .957 & .947 & .929 & \textBF{.974} &.913&.942&.952&\underline{.967}&\textBF{.976}&.936&.952 &\textBF{.963}&\underline{.931}&.849&.861\\
Pneumothorax && .960 & .975 & \textBF{.992} & \underline{.988} & .973 & .944 & .956 &.894&.936&.942&\underline{.959}&\textBF{.979}&.926& .951  &.711&.898&\underline{.961}&\textBF{.963}\\
Pleural Effusion && .968 & .917 & \underline{.974} & .962 & \textBF{.976} & .943 & .959 &.896&.921&.933&\underline{.945}&\textBF{.981}&.923& .941&.891&.937&\underline{.967}&\textBF{.968}\\
Pleural Other&& .939 & .945 & \underline{.985} & .969 & .958 & \textBF{.971} & .961 &.893&.904&.914&.925&\textBF{.964}&.921&\underline{.930}&.872&.893&\underline{.931}&\textBF{.939}\\
Fracture&& .926 & .860 & .892 & .893 & .938 & .935 & \textBF{.942} &.832&.886&.898&.906&.958&\underline{.975} & \textBF{.978}&.861&.875&\underline{.896}&\textBF{.907}\\
Support Devices  && \underline{.904} & .813 & .847 & .872 & \textBF{.912} & .899 & .903 &.877&.919&.921&.934&\textBF{.983} &.925&\underline{.941}&\underline{.924}&\textBF{.928}&.860&.889\\ \cmidrule(r){1-2}\cmidrule(rl){3-9}\cmidrule(rl){10-16}\cmidrule(l){17-20}
Average AUC  && .959 & .930 & .930 & \underline{.965} & \textBF{.970} & .947 & \underline{.965} &.896 & .919&.921&\underline{.947}&   \textBF{.974} &.931 & .945&.882&.919&\underline{.951}&\textBF{.967}\\\cmidrule(r){1-2}\cmidrule(rl){3-9}\cmidrule(rl){10-16}\cmidrule(l){17-20}
Recall      &  & .817 & .769 & .819 & .818 & \underline{.863} & \textBF{.867} & \underline{.863} & .827 & .743 & .841 & .847 & \textBF{.868}                           & .842 & \underline{.848} & .510 & .679 & \underline{.823} & \textBF{.841}                                                                                                                                                                               \\
Precision   &  & .607 & .561 & .572 & .602 & \textBF{.629} & .615 & \underline{.624} & .526 & .525 & .594 & .601 & .611                           & \underline{.614} & \textBF{.631} & .220 & .491 & \underline{.544} & \textBF{.587}                                                                                                                                                                               \\
F1          &  & .719 & .681 & .718 & .700 & \textBF{.721} & \underline{.720} & .716 & .603 & .615 & .689 & .714 & \textBF{.726}                           & .713 & \underline{.718} & .556 & .575 & \underline{.656} & \textBF{.681}\\\bottomrule
\end{tabular}
}

\label{tab:clsr}
\end{table}
\section{Results \& Discussion}
\subsubsection{Report classification}\autoref{tab:clsr} shows disease classification results of our graph embeddings. Results are shown for an encoder that performed best on average with hidden size $1024$ and three graph attention layers. Our method yields competitive performance compared to BioBERT and PubMedBERT, and slightly lower compared to ClinicalBERT, while being $200 \times$ smaller in terms of parameters (\autoref{fig:graph_param}). On the Spanish PadChest dataset our method outperforms BERT-based methods. This can be attributed to the size of the training corpora of these models, which is ten times smaller than their English counterparts. Next to this, both CPU and GPU inference rates (samples per second) are faster for our method. Our method performs relatively better on the smaller OpenI dataset, emphasising the effectiveness of our embedding for report representation without relying on large datasets. 
\begin{table}[!t]
\setlength\tabcolsep{1em}
\centering
\caption{Comparison of disease classification metrics on variational knowledge distillation (VKD) based chest X-ray classification. Performance is reported on single-modal image classification, VKD with ClinicalBERT/Bio-cli-52k report embeddings and VKD with our proposed graph embeddings. }
\resizebox{\textwidth}{!}{%
\begin{tabular}{lccccccccc}
\toprule
                           \multicolumn{1}{c}{}& \multicolumn{3}{c}{OpenI}                                                & \multicolumn{3}{c}{MIMIC-CXR} &\multicolumn{3}{c}{PadChest}                                \\ \cmidrule(r){2-4}\cmidrule(rl){5-7}\cmidrule(l){8-10}
                           \multicolumn{1}{c}{}&  Img only &  ClinicalBERT&  Ours& Img only &  ClinicalBERT& Ours&Img only &  Bio-cli-52k& Ours\\\cmidrule(r){1-1}\cmidrule(rl){2-4}\cmidrule(l){5-7}\cmidrule(l){8-10}

Recall&.568 & .582 & \textBF{.578} & .546 & \textBF{.565} & .526 & .519 & .529 & \textBF{.533} \\
Precision&.487 & \textBF{.534} & .527 & .530 & \textBF{.538} & .536 & .487 & .510 & \textBF{.530} \\
F1&.491 & \textBF{.510} & .495 & .468 & \textBF{.509} & .487 & .468 & \textBF{.481} & .473\\\cmidrule(r){1-1}\cmidrule(rl){2-4}\cmidrule(l){5-7}\cmidrule(l){8-10}
AUC          & .837                           & \textbf{.885} & .862 & .807                           & \textbf{.839} & .823 &.802&.815&\textBF{.819}\\
\bottomrule
\end{tabular}
}

\label{tab:comparison}
\end{table}

\subsubsection{Cross-modal knowledge transfer}

The application of graphs in a multi-modal setting can give a better understanding on how well the graph captures complex information structures that can pass across modalities. \autoref{tab:comparison} shows the results of our method on disease classification of chest X-rays with cross modal knowledge transfer, compared to existing methods using ClinicalBERT as report embedding. Training of this framework showed that convergence of this model is complex for graphs with shallow encoders with hidden layers of smaller size. \autoref{tab:comparison} thus shows results with an encoder of 12 graph attention layers and hidden size 2048. 

While not outperforming the existing ClinicalBERT method, graph embeddings show to work on both MIMIC-CXR and OpenI. There is increased performance compared to image-only classification so we can successfully transfer information across modalities without requiring large pre-training datasets. 


\subsubsection{Ablation studies}

We analyze graph encoders for disease classification in Figure \ref{fig:graph_param}, showing the effect of encoder count and hidden size on performance. Parameter count is crucial, with ClinicalBERT performing best but requiring more resources. The performance difference between the smallest (0.4M parameters) and largest (62M parameters) models is small, indicating that graph construction captures medical knowledge well regardless of encoder size. We also conduct an ablation on the components of the graph in \autoref{tab:le}. The role of node types is shown by removing them from the graph. Key elements in the graph structure appear to be the global node and edges between SNOMED CT concepts. This accentuates how the combined report composition and the incorporation of medical knowledge base SNOMED CT creates a rich representation of the report.

        
    

\begin{table}[!b]
	\begin{minipage}[b]{0.60\linewidth}
		
		\raggedleft
		\includegraphics[width =\textwidth]{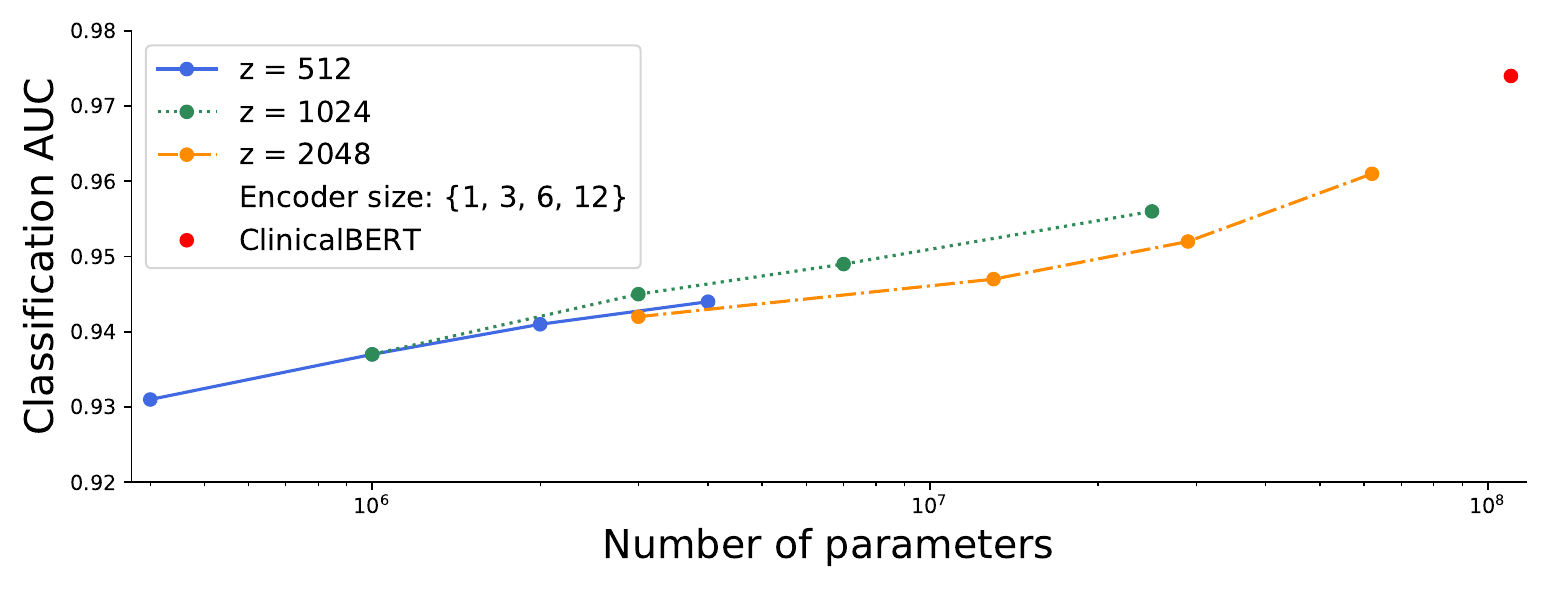}
        \captionof{figure}{Graph size and performance for a set of models with hidden size $z$ and varying encoder size.}
        \label{fig:graph_param}
	\end{minipage}\hfill
	\begin{minipage}[b]{0.37\linewidth}
		\RaggedRight
		\resizebox{\textwidth}{!}{%
		\begin{tabular}{lccc}
\toprule
&OpenI&MIMIC-CXR&PadChest   \\\cmidrule(rl){2-2}\cmidrule(rl){3-3}\cmidrule(rl){4-4}
Full graph                                                      &\textBF{0.947}&\textBF{0.931}&\textBF{0.951} \\ 
w/o $\mathbf{g}$                                            &0.940&0.925&0.942 \\ 
w/o $\mathbf{g} \& \mathbf{s}$                          &0.932&0.918&0.934\\ 
w/o $e^{c \leftrightarrow c}$                                   &0.936&0.918&0.926 \\ 
w/o $\mathbf{g} \& \mathbf{s} \& e^{c \leftrightarrow c}$   &0.916&0.920&0.921 \\ \bottomrule
\vspace{8mm}

\end{tabular}}

    \caption{Graph ablations on disease classification.}
    \label{tab:le}
		
	\end{minipage}
\end{table}


\begin{figure}[!t]
    \centering
    \includegraphics[width = 0.90\linewidth]{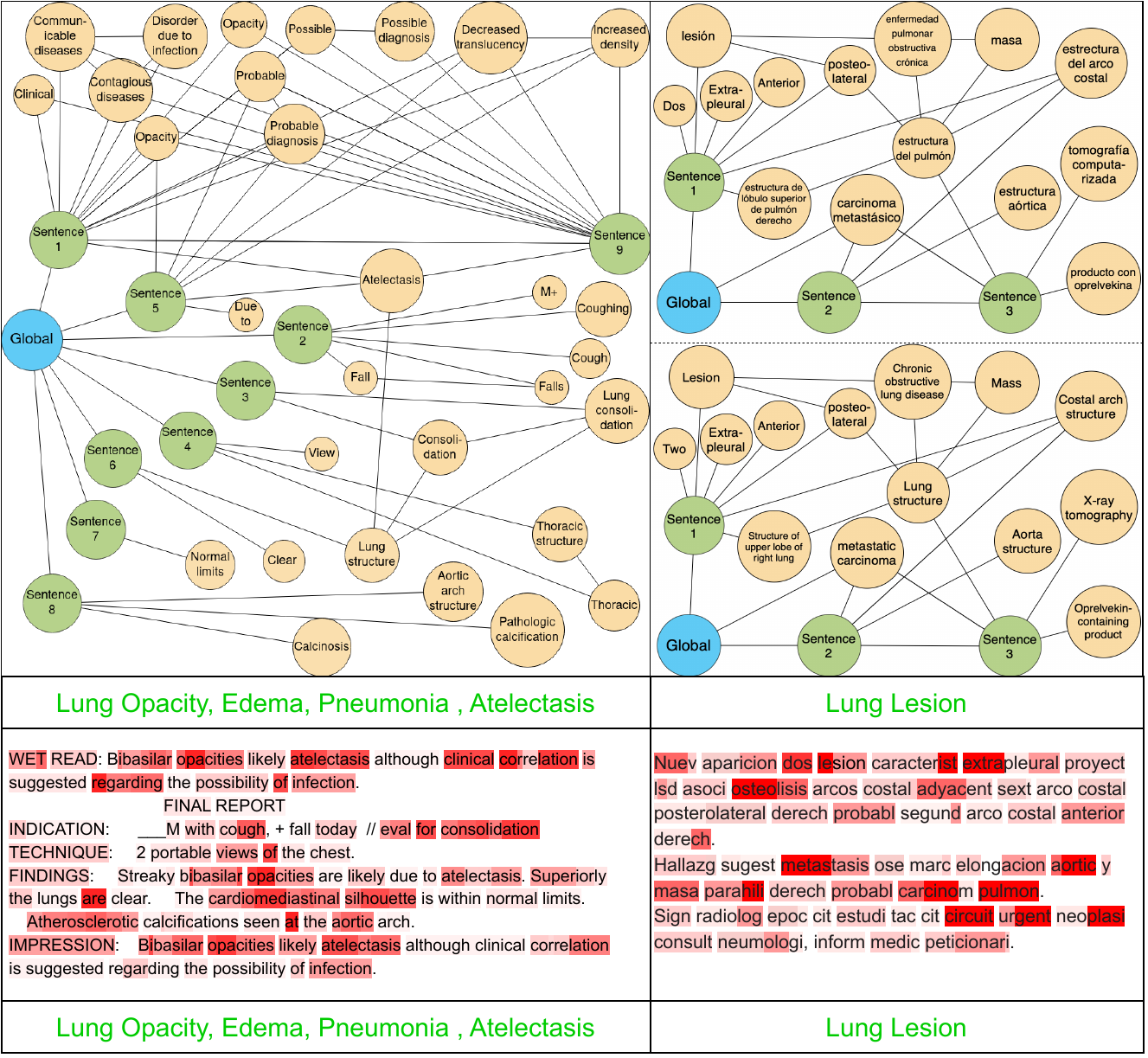}
    \caption{Examples of our graph embedding (top), compared to ClinicalBERT embedding weights (bottom), with predicted disease labels. Green means correct classification and red false negative. Global node edges are omitted for clarity.}
    \label{fig:3examples}
\end{figure}
\subsubsection{Graph visualization and explainability}
\autoref{fig:3examples} shows two examples of our graph embedding and attention weights of ClinicalBERT with results of these methods on disease classification. The left report shows how disease labels are verbatim present in the detected SNOMED CT concepts, contributing to interpretability. Concepts that reoccur in the report or have a high `connectedness' are often important. The graph handles repeated terms more efficiently than clinicalBERT: our graph consists of 34 nodes, while tokenization with ClinicalBERT takes as much as 124 tokens. Our report graph captures the word `opacities' in its entirety. In ClinicalBERT, this is tokenized as `o', `pa' and `cities'. The last token in this sequence obviously has another contextualized meaning in the general language BERT model on which ClinicalBERT is based. This illustrates graphs can captures medical terminology in a more intuitive and interpretable way. 

The last example shows a Spanish report graph, and below it the English translation. Translation is straightforward since each content node corresponds to a multi-lingual UMLS concept.   


\section{Conclusion}
In this paper, we presented a knowledge graph based method for structured representations of radiology reports. The knowledge graph embeddings explicitly encode medical knowledge from clinical knowledge bases for transfer across domains and languages without the heavy reliance on large amounts of data. Meanwhile we also keep the model size needed for training magnitudes smaller than existing BERT-based models.  The proposed graph based representations can yield comparable results to current state-of-the-art Transformer-based models on English and Spanish language by capturing both structural relations and content relations found in existing knowledge bases, resulting in more informative representations of radiology reports
\section*{Acknowledgements}
This work is financially supported by the Inception Institute of Artificial Intelligence, the University of Amsterdam and the allowance Top consortia for Knowledge and Innovation (TKIs) from the Netherlands Ministry of Economic Affairs and Climate Policy.
\bibliographystyle{splncs04}
\bibliography{sources}

%




\end{document}